\documentclass[
]{ceurart}

\usepackage{listings}
\usepackage{graphicx}
\usepackage{amsmath} 
\usepackage{amssymb}  
\usepackage{subcaption}
\usepackage{booktabs}
\usepackage{fvextra}
\usepackage{xcolor}

\usepackage{tcolorbox}

\begin{document}

\copyrightyear{2025}
\copyrightclause{Copyright for this paper by its authors.
  Use permitted under Creative Commons License Attribution 4.0
  International (CC BY 4.0).}


\title{Conversational Code Generation: a Case Study of Designing a Dialogue System for Generating Driving Scenarios for Testing Autonomous Vehicles}

\author{Rimvydas Rubavicius}[%
orcid=0000-0003-4270-7154,
email=rimvydas.rubavicius@ed.ac.uk,]
\cormark[1]
\address{School of Informatics, University of Edinburgh 10 Crichton
Street, Edinburgh EH8 9AB}

\author{Antonio Valerio Miceli-Barone}[%
orcid=0000-0002-9904-8869,
email=amiceli@ed.ac.uk,
]

\author{Alex Lascarides}[%
orcid=0000-0003-1704-1864,
email=alex@inf.ed.ac.uk,
]

\author{Subramanian Ramamoorthy}[%
orcid=0000-0002-6300-5103,
email=s.ramamoorthy@ed.ac.uk,
]

\cortext[1]{Corresponding author.}

\conference{GeCoIn 2025: Generative Code Intelligence Workshop, co-located with the 28th European Conference on Artificial Intelligence (ECAI-2025),  October 26, 2025 --- Bologna, Italy}

\begin{abstract}
Cyber-physical systems like autonomous vehicles are tested in simulation before deployment, using domain-specific programs for scenario specification. To aid the testing of autonomous vehicles in simulation, we design a natural language interface, using an instruction-following large language model (LLM), to assist a non-coding domain expert in synthesising the desired scenarios and vehicle behaviours. We show that using it to convert utterances to the symbolic program is feasible, despite the very small training dataset. Human experiments show that \textit{dialogue} is critical to successful simulation generation, leading to a 4.5 times higher success rate than generation without engaging in extended conversation.
\end{abstract}

\begin{keywords}
  Code Generation \sep Autonomous Vehicles \sep Simulation \sep Human-computer Interaction 
\end{keywords}

\maketitle

\section{Introduction}
\label{sec:introduction}

Testing autonomous vehicles exclusively on public roads, especially in near-crash scenarios, is not feasible~\cite{KALRA2016182}. Iterative development and testing in simulation is essential~\cite{10.1007/978-3-658-21194-3_82}, especially in the early stages of the development process. Simulators like CARLA~\cite{DBLP:conf/corl/DosovitskiyRCLK17} support this iterative development process for autonomous vehicles (AVs). Using CARLA, engineers can generate driving scenarios using Scenic~\cite{DBLP:journals/ml/FremontKDGYSS23}: probabilistic scenario description languages such as Scenic specify a distribution over the driving scenarios that satisfy the description, and then sample many simulation instances for evaluation. In this way, challenging driving scenarios can be generated~\cite{DBLP:journals/pami/LiPFZXZ23,li2023scenarionet}, which can then be used for evaluation of control algorithms for autonomous vehicles without a physical system, safely and cost-effectively. 

Writing programs to specify driving scenarios in Scenic, or more generally programming in domain-specific languages (DSLs), is challenging with steep learning curves. Domain experts with deep knowledge of edge cases and desired driving behaviours may lack proficiency in Scenic or other DSLs, or even when they do have such proficiency, they may still benefit from some assistance in their workflow. A natural language interface in the form of a dialogue system in which engineers converse with a chatbot that knows how to program in Scenic is a useful tool for facilitating this. In this paper, we aim to develop such a dialogue system, and by doing so increase engineer access to simulation-driven testing, by allowing them to interactively synthesise scenarios using natural language dialogue. 

\begin{figure}
    \centering
    \includegraphics[width=\columnwidth]{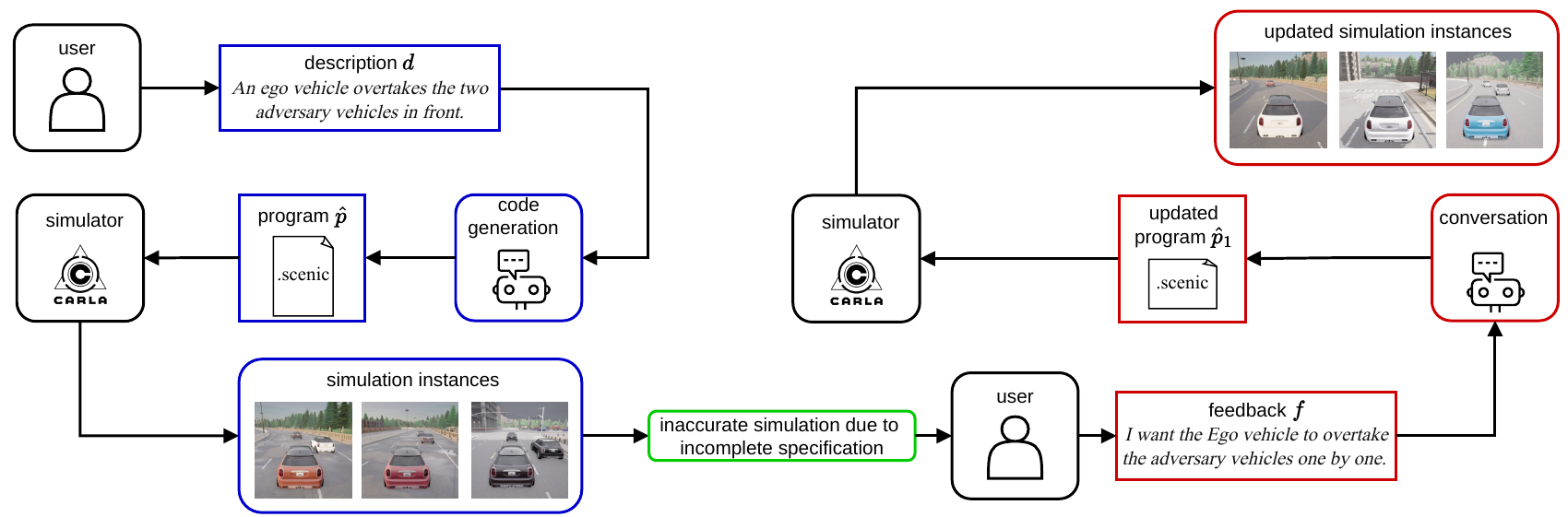}
    \caption{Dialogue System Overview. It facilitates driving scenario generation using natural language in code generation (\S~\ref{sec:code_generation}) and  conversation (\S~\ref{sec:conversation}).}
    \label{fig:overview}
\end{figure}

This is a challenging task for two reasons. Firstly due to data scarcity (only 32 examples of Scenic programs with natural language descriptions are available online), approaches in developing a dialogue system using instruction-following large language models (IFLLM)~\cite{NEURIPS2022_b1efde53} are not directly applicable. Secondly, the scenario description provided by the user may be {\em underspecified}: missing some details about the desired driving scenario in the description that is only exposed through interaction with the simulator. Figure~\ref{fig:overview} illustrates this phenomenon. When a user utters ``an ego vehicle overtakes the two adversary vehicles'', the dialogue system can synthesize the simulation instance satisfying the description, but the description itself does not provide all information about the situation, which in this case is the manner of overtaking (could be one by one, or both at the same time). The user only becomes aware of the necessity to provide such details by observing simulation instances.  In this context, it would be helpful for a natural language interface to allow the user to express feedback to the system, which can then be used to update the program and, in turn, generate simulation instances.

To this end, this paper makes the following contributions: 1) we create a dataset of English description-Scenic program pairs for a variety of driving scenarios; 2) we investigate how IFLLM can be used to synthesize Scenic programs from natural language expressions in an embodied conversation where the user converses with a chatbot and reacts to driving simulation instances; and 3) we conduct human trials to evaluate the dialogue system, and in particular the value of having multiple dialogue turns. Our results demonstrate that the dialogue system can be a valuable tool for facilitating the generation of interactive driving scenarios.

\section{Related Work}
\label{sec:related_work}

\paragraph{Code Generation} The problem of converting natural language to executable programs has been widely studied as semantic parsing~\cite{wong-mooney-2006-learning,zettlemoyer-collins-2009-learning,dong-lapata-2016-language}. Recently, using IFLLMs has been explored for this task and also the wider problem of arbitrary code generation, including auto-completion~\cite{DBLP:journals/corr/abs-2406-00515} to assist the software development process. In this paper, we concentrate on developing a dialogue system that generates Scenic programs using LLMs. It is a low-resource domain where only a handful of examples are available. This is in contrast to other popular programming languages like Python, which benefit from many code repositories for training~\citep{DBLP:journals/tmlr/KocetkovLALMJMF23}. Furthermore, the code for high-resource programming languages follows style guidelines, making it easier for models to learn generalisations from the well-structured data.  In contrast, the Scenic program repository was written in different styles by different software engineers.

\paragraph{Testing AVs} Testing is a fundamental process within software engineering, which is particularly important for designing safety-critical systems, such as AVs~\cite{Feng2021,Padmaja2023}. There are a variety of methods for testing and validating AV design, even with black-box components~\cite{DBLP:journals/jair/CorsoMKLK21}. This paper focuses on testing by generating driving scenarios as a complementary procedure to formal verification. Such testing involves interaction with expert users, which is essential for human-centred design~\cite{Rosenbrock1989DesigningHT}. Driving scenario generation has been explored before: both direct scenario generation~\cite{10529537,10160296} or by utilising Scenic~\cite{miceli-barone-etal-2023-dialogue, zhang2024chatscene}. What is unique about this work is: (i) human experiments; and (ii) utilising dialogue to align the generated driving scenarios with the user's intent.  We view this latter contribution as essential, given the ubiquitous phenomenon of {\em natural language pragmatics} \cite{grice:1969}: speakers of natural language often intend to convey a meaning that goes beyond what they make linguistically explicit, relying instead on their interlocutor's capacity to decode the intended content using the linguistic and non-linguistic context.  There is no guarantee that the sampling processes in Scenic match how competent interlocutors perform this task, and so errors in decoding the intended message via Scenic sampling may happen, prompting the need for user feedback.

\paragraph{Dialogue Systems and Tools} Natural language interfaces in the form of a dialogue system have become a popular means for users to interact with systems without needing to know the underlying system or how to program it\citep{10.1145/3166054.3166058,Ni2023}. Broadly, dialogue systems are categorised by their function as goal-oriented or for chit-chat. We are designing a goal-oriented dialogue system with tool use~\citep{NEURIPS2023_d842425e}, which in this case is the CARLA simulator. What is unique about our work is that the driving scenarios generated by the simulator are not observed in the dialogue system and are latent, with observations coming only from the user's feedback after observing simulation instances. 

\section{The Dialogue System}
\label{sec:dialogue_system}

\begin{figure*}
\begin{subfigure}{.5\textwidth}
  \centering
  \includegraphics[width=\linewidth]{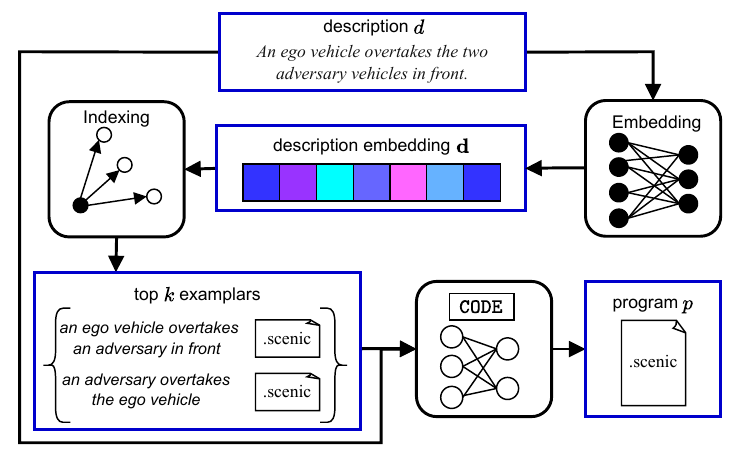}
  \caption{Code generation using RAG \cite{NEURIPS2020_6b493230} (\S~\ref{sec:code_generation})}
  \label{fig:code_generation_mode}
\end{subfigure}%
\begin{subfigure}{.5\textwidth}
  \centering
  \includegraphics[width=\linewidth]{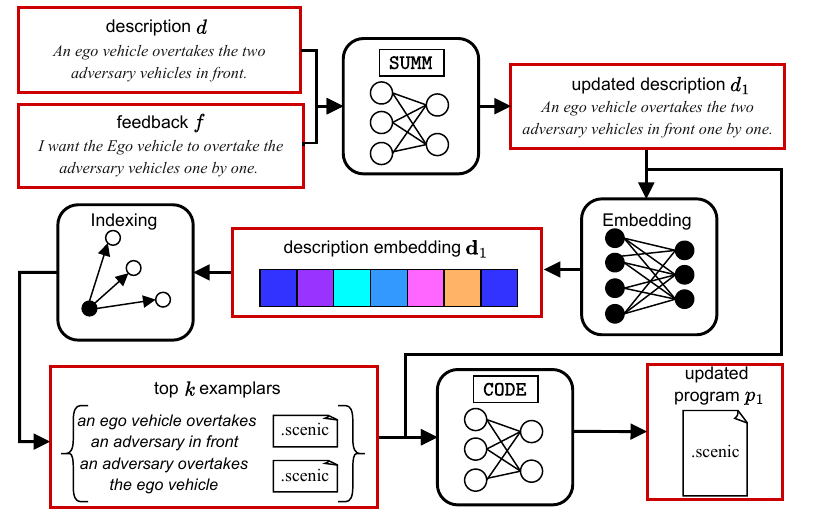}
  \caption{Conversation for updating description \(d_1\) (\S~\ref{sec:conversation}).}
  \label{fig:conversation_model}
\end{subfigure}
\caption{Dialogue System for Conversational Driving Scenario Generation.}
\label{fig:dialogue_system}
\end{figure*}

This section describes the dialogue system for generating driving simulations via conversations. Figure~\ref{fig:dialogue_system} demonstrates the two modes of the system: code generation (\S\ref{sec:code_generation}); and conversation (\S\ref{sec:conversation}), in which the user responds to simulation instances as depicted in Figure~\ref{fig:overview}.

\subsection{Code Generation}
\label{sec:code_generation}

\begin{figure}
    \centering
    \begin{tcolorbox}[title=\(\mathtt{CODE}\) prompt for code generation.]
        [INST] You are a helpful assistant that translates English descriptions to Scenic programs. Scenic is a domain-specific probabilistic programming language for creating distributions over specified scenarios. For driving scenarios, each program has the following blocks:
        \begin{itemize}
            \item MAP AND MODEL: importing town assets and enabling simulator; 
            \item CONSTANTS: specifying vehicle blueprint and other constants like vehicle speed, brake intensity and safety distance;
            \item AGENT'S BEHAVIOUR: describing how individual vehicles behave in the scenario;
            \item SPATIAL RELATIONS: outlining the type of road the scenario needs to be synthesized in (e.g. having or not having intersections)
            \item SCENARIO SPECIFICATION: creating individual vehicles and pedestrians in the specified roads, together with constraints that are required to be true for the full simulation as well as the termination condition. 
        \end{itemize}
        Here are examples of descriptions and programs: \{DESCRIPTION-PROGRAM PAIRS\}
        
        Now, please translate the following English description to a Scenic program. Just give the program. No extra information. Description: \{DESCRIPTION\}[/INST]
    \end{tcolorbox}
    \begin{tcolorbox}[title=\(\mathtt{SUMM}\) prompt for description update.]
        [INST] You are a helpful assistant that creates updated driving scenario descriptions. Given a driving scenario description and corresponding feedback regarding the simulation, your task is to generate an updated description that incorporates the feedback. Please ensure that the updated description accurately reflects the intended action based on the feedback received and does not introduce additional information or lose information, like the number of vehicles or pedestrians in the situation. The description will outline a specific scenario or context, while the feedback will provide information about how the described scenario could have been improved or modified. Be sure to maintain the original meaning and intent of both the initial description and the feedback. Be brief and only return the updated driving scenario description.
        Description: \{DESCRIPTION\}
        Feedback: \{FEEDBACK\}[/INST]
    \end{tcolorbox}

    \caption{Interactive driving simulation generation prompts. Curly brackets denote task-specific information.}
    \label{fig:prompts}
\end{figure}

To generate Scenic programs from natural language descriptions, the dialogue system uses retrieval-augmented generation (RAG) \cite{NEURIPS2020_6b493230} with in-context learning \cite{NEURIPS2020_1457c0d6} to construct the \(\mathtt{CODE}\) prompt.

To generate a program \(\hat{p}\) from natural language description \(d\), we constructed a dataset \(\mathcal{D} =\{(d_i,p_i)\}_{i=1}^{N}\) of size \(N=105\) description-program pairs\footnote{\url{https://huggingface.co/datasets/assistive-autonomy/scenic-driving-scenarios}}, via manual augmentation of the few exemplars available on the web.\footnote{\url{https://github.com/BerkeleyLearnVerify/Scenic/tree/2.x/examples}} Additionally, in our manual augmentation process, we standardise the codebase to minimise variability and ease the learning process in this low-resource regime. Each program is standardised in its coding style, with five distinct code blocks---map and model, constants, agent's behaviour, spatial relations, and scenario specification---to minimise the variation between the exemplars. The dataset \(\mathcal{D}\) 
is indexed via embeddings of descriptions.\footnote{\href{https://huggingface.co/BAAI/bge-small-en-v1.5}{https://huggingface.co/BAAI/bge-small-en-v1.5}}

At inference time, the embedding \(\mathbf{d}\) of the user's description \(d\) is used to find the most similar \(k=3\) description-program pairs from $\mathcal{D}$ using maximum inner-product search:
\begin{equation}
    \mathcal{D}_k(\mathbf{d}) = \{ (d,p)\in\mathcal{D} \mid (d,p) \in top(k,\mathbf{d}, \mathcal{D})\} 
\end{equation}
The description with the retrieved exemplars are then used to create the \(\mathtt{CODE}\) prompt (see Figure~\ref{fig:prompts} for the prompt specification) which is then used to generate the program \(\hat{p}\) using an \(\mathtt{IFLLM}\):
\begin{equation}
    \hat{p} = \mathtt{IFLLM}(\mathtt{CODE}(\mathcal{D}_k(\mathbf{d}), d))
\end{equation}
There is no guarantee that the program \(\hat{p}\) will execute. To fix this, we perform error feeding: the program \(\hat{p}\) is passed to the simulator to attempt to create simulation instances; and if an error occurs, the natural language description, $\hat{p}$ and its errors are passed back to the \(\mathtt{IFLLM}\), to attempt to self-correct the output. If, after three attempts of error feeding, the generated program doesn't execute, the user is asked to provide an alternative description.

\subsection{Conversation}
\label{sec:conversation}
When $\hat{p}$ executes, \(n=3\) instances of the driving scenario simulations are shown to the user, who is given the opportunity to react to these, either by indicating successful generation or by providing natural language feedback \(f\) that corrects or refines the results, in an attempt to provide simulations that better align with the user's communicative intent.

Feedback is a response to the \textit{simulations} and not to \(\hat{p}\) (which is not observable to the user): it can feature and refer to information that is not present in the Scenic program but is present in the simulation due to the nature of random sampling. Because of this, feeding only $f$ to \(\mathtt{IFLLM}\) is suboptimal due to its missing context, which cannot be given in a principled way (e.g. conditioning on the simulation video or some domain-specific trace would lead to further data scarcity, which is already problematic in this domain). To alleviate this issue, we feedback an updated description by summarising $d$ and $f$, using a \(\mathtt{SUMM}\) prompt (see the Figure~\ref{fig:prompts} for the prompt specification), to generate the updated description \(d_1\). With this description, our aim is to guide \(\mathtt{IFLLM}\), minimise the overall context, and not introduce hallucinations and unwanted behaviours. 
\begin{equation}
        d_1 = \mathtt{IFLLM}(\mathtt{SUMM}(d,f)) 
\end{equation}
The updated description \(d_1\) is used to generate the updated program \(\hat{p}_1\) following the previously described code generation procedure.
\begin{equation}
    \hat{p}_1 = \mathtt{IFLLM}(\mathtt{CODE}(\mathcal{D}_k(\mathbf{d}_1), d_1))
\end{equation}
The conversation continues up to \(m=4\) turns.  If the user remains unsatisfied with the simulations, the conversation is deemed unsuccessful.

\section{Experiments}
\label{sec:experiments}
We conduct experiments for code generation and human evaluation for the overall dialogue system. NVIDIA RTX A6000 is used for serving IFLLM via a text generation inference server. For human trials, we use an Alienware Area-51m laptop with NVIDIA RGX 2080 to generate simulation instances in CARLA.

\subsection{Code Generation Experiments}
\label{sec:code_generation_experiments}

\subsubsection{Experimental Setup}
We experiment with several open-source IFLLMs with 7B parameters: Mistral~\cite{DBLP:journals/corr/abs-2310-06825}, Gemma~\cite{DBLP:journals/corr/abs-2403-08295}, and CodeLlama~\cite{DBLP:journals/corr/abs-2308-12950} that used \(k=3\) exemplars to construct \(\mathtt{CODE}\) that's retrieved using RAG, or at random with and without error feeding. To cope with the small dataset bias when evaluating different IFLLMs, we perform leave-one-out validation~\cite{DBLP:books/lib/Bishop07}. We record three metrics: (a) BLEU~\cite{papineni-etal-2002-bleu,post-2018-call} to measure generation precision; (b) ROUGE-L~\cite{lin-2004-rouge} to measure generation recall; and (c) Execution (EXEC) to measure the percentage of the generated code that is executable, all measured with error feeding (if required). More sophisticated code generation evaluation metrics like CodeBLEU~\cite{DBLP:journals/corr/abs-2009-10297}, to abstract away from the surface form and focus on the syntactic similarity, are not considered because such metrics require adaptation of the parser to Scenic programs. This is partially circumvented by program standardisation performed in the dataset construction.

\subsubsection{Results and Discussion}

Table~\ref{tab:code_geneation} records the results for code generation. We observe that IFLLM that uses \(\mathtt{CODE}\) prompt constructed using RAG is better than a random selection, which most of the time results in non-executable programs. Error feeding boosts performance for IFLMM that are not fine-tuned on code (Mistral and Gemma ), yet it does increase inference time, which for Gemma leads to non-termination in a reasonable time. High BLEU and ROUGE-L indicate a high correspondence between prediction and reference, as expected after program standardisation. For EXEC, CodeLlama is better than Mistral and Gemma, which is as expected due to the supervised fine-tuning on code. Based on code generation experiments, we chose CodeLlama with RAG and error feeding to use for human evaluation.

\begin{table}[]
\caption{Code generation evaluation. RAG notes that \(\mathtt{CODE}\) prompt was constructed using RAG while RAND notes it was constructed using random exemplars. EF notes error feedback while symbol - denotes cases when IFLLM did not terminate in a reasonable time (more than 10 min.).}
\label{tab:code_geneation}
\resizebox{\columnwidth}{!}{%
\begin{tabular}{@{}llll|llll|llll@{}}
\toprule
{CodeLlama}  & BLEU           & ROUGE-L        & EXEC             & Gemma & BLEU           & ROUGE-L        & EXEC                     & Mistral & BLEU           & ROUGE-L        & EXEC    \\ \midrule
RAG+EF        & 64.32          & 63.23          & \textbf{71.43} & RAG+EF    & 56.86                 & 55.86                 & 40.95                  & RAG+EF      & 84.54 & 79.94   & 41.90 \\
RAG              & \textbf{89.66} & \textbf{85.96} & 70.48          & RAG          & 10.44                 & 16.79                 & 0.00                   & RAG            & 87.80 & 84.03   & 37.14 \\
RAND+EF     & 0.09           & 5.80           & 0.00           & RAND+EF & \multicolumn{1}{c}{-} & \multicolumn{1}{c}{-} & \multicolumn{1}{c|}{-} & RAND+EF   & 10.41 & 30.74   & 1.90  \\
RAND           & 0.45           & 17.29          & 0.00           & RAND       & \multicolumn{1}{c}{-} & \multicolumn{1}{c}{-} & \multicolumn{1}{c|}{-} & RAND         & 4.46  & 23.05   & 0.00  \\ \bottomrule
\end{tabular}
}
\end{table}

\subsection{Human Evaluation}
\label{sec:human_evaluation}

\begin{figure*}
    \centering
    \includegraphics[width=\linewidth]{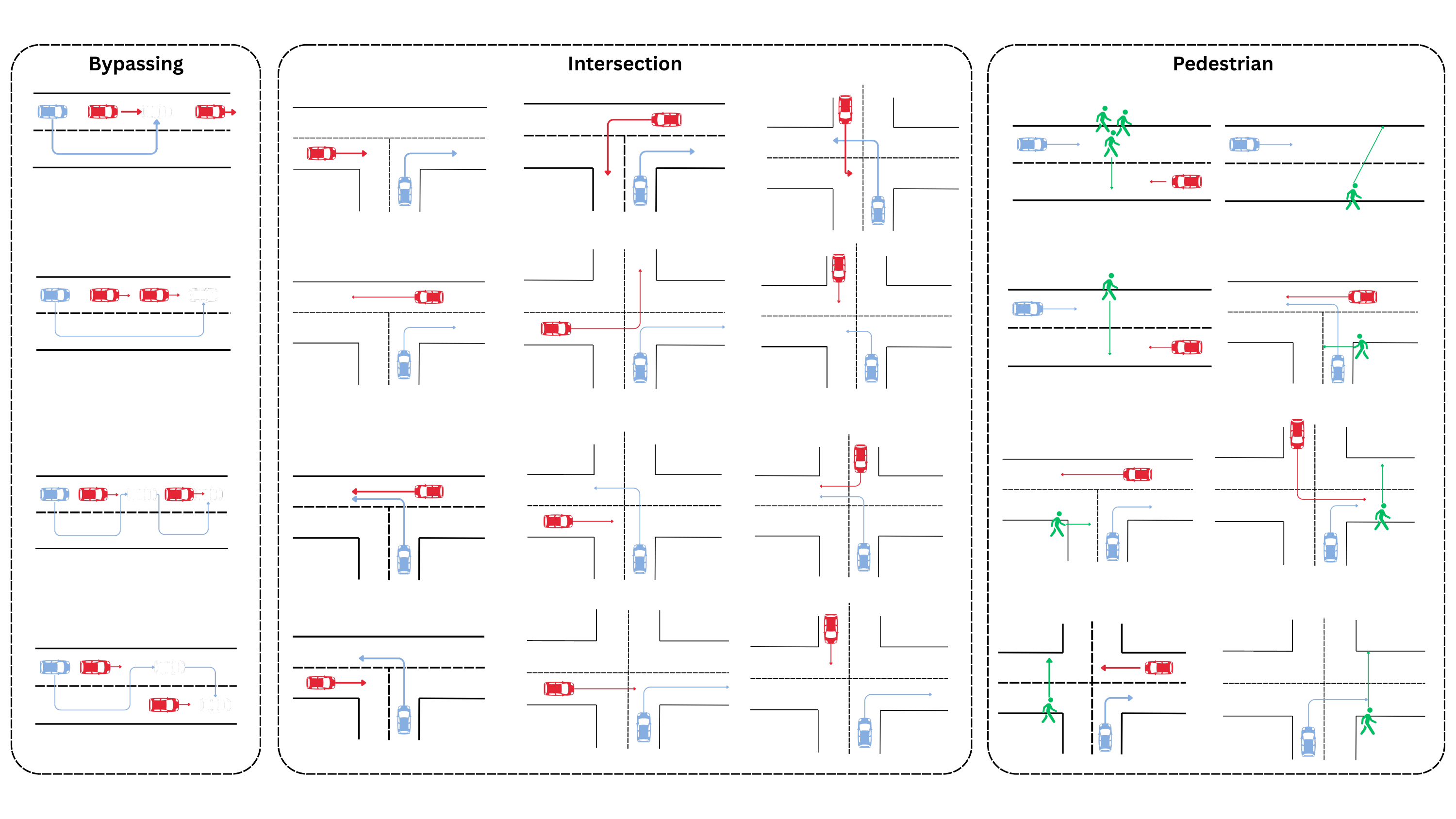}
    \caption{Visual stimuli for generating driving scenarios using dialogue system in human evaluation. In each stimulus, blue elements denote an ego vehicle, red elements denote adversary vehicles, and green elements denote pedestrians.}
    \label{fig:stimuli}
\end{figure*}

\begin{figure*}
\begin{subfigure}{.5\linewidth}
  \centering
  \includegraphics[width=\linewidth]{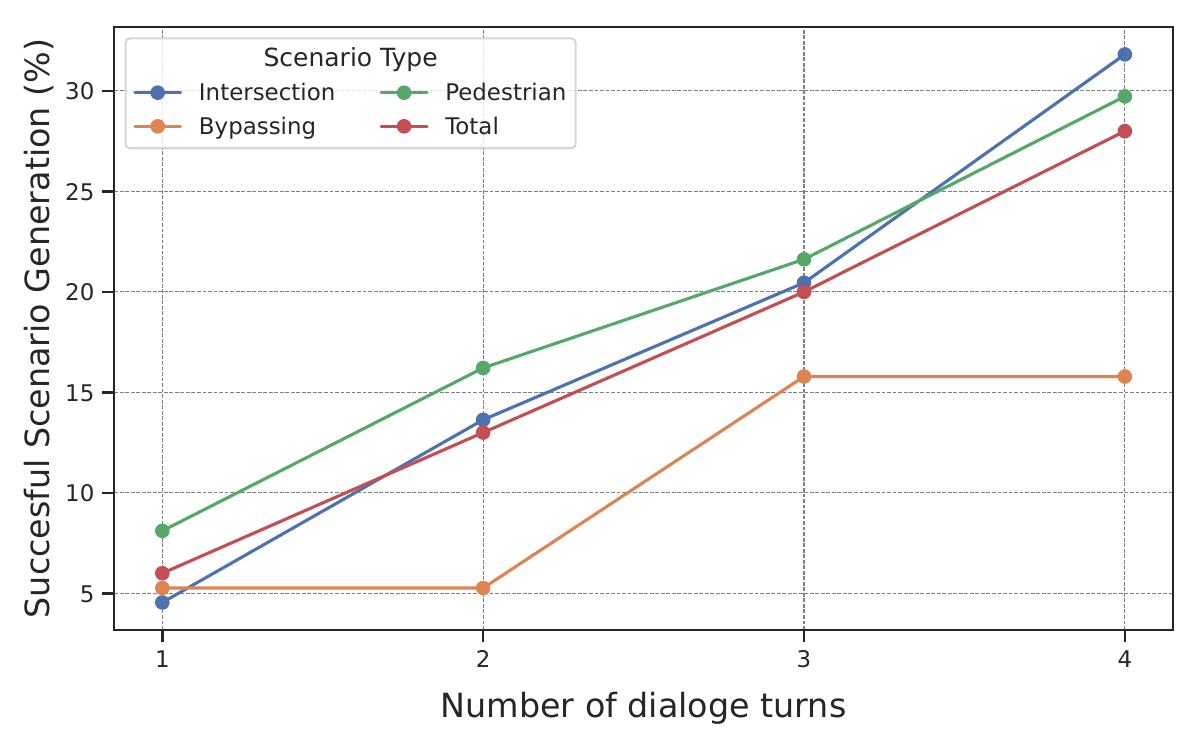}
  \caption{With execution errors}
  \label{fig:results_1}
\end{subfigure}%
\begin{subfigure}{.5\linewidth}
  \centering
  \includegraphics[width=\linewidth]{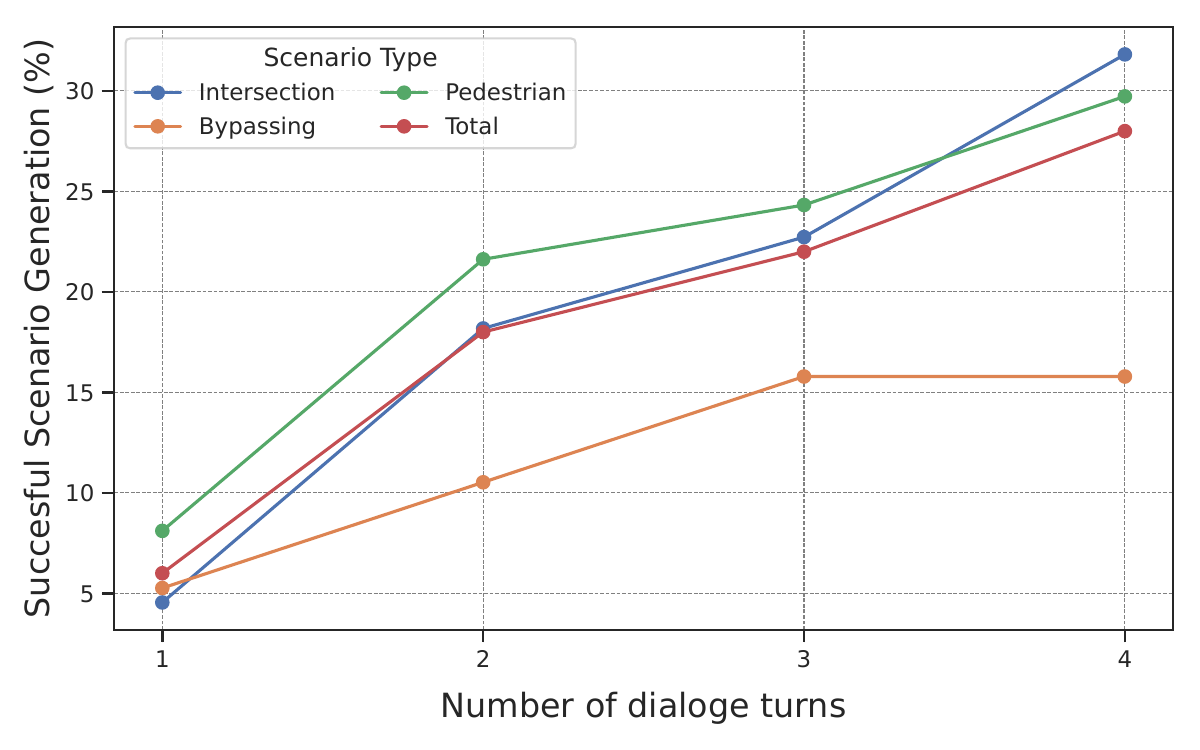}
  \caption{without execution errors}
  \label{fig:results_2}
\end{subfigure}
\caption{Commutative scenario generation success over the number of dialogue turns in human evaluation with (\ref{fig:results_1}) and without (\ref{fig:results_2}) execution errors. For example, if, on the 3rd turn, the system fails to produce executable code and the user is asked to paraphrase, leading to successful scenario generation, we count this as success on 4 dialogue turns with execution errors and 3 dialogue turns without execution errors. The results show a positive correlation between the successful scenario generation and the number of dialogue turns for all types of scenarios considered.
}
\label{fig:results}
\end{figure*}

\subsubsection{Experimental Setup}

We perform a human evaluation of the dialogue system. Visual stimuli used for human trials are given in Figure~\ref{fig:stimuli}.  The user is presented with a (diagrammatic) stimulus and asked to describe it. After that, CARLA is run to produce 3 simulation instances.\footnote{The dialogue system can produce many (different) instances, but to keep the stimuli contained, we limit it to 3.} After observing it, the user may respond with further natural language input to what they see, or check ``satisfied with the scenario produced'' and move to the next stimulus. The user can have up to \(m=4\) dialogue turns. If code generation does not produce executable code, the user is asked to paraphrase their description as if it is a fresh interaction. The stimuli are categorised into three types: bypassing (e.g. the user stimulus may yield the user description ``an ego overtakes an adversary''), intersections (e.g. ``an ego goes left at a 4-way intersection while the adversary approaches from behind.''), and pedestrians (e.g.``an ego goes left and yields to the pedestrian''). The experiments were conducted with 20 users who have a driving license (as a proxy for domain proficiency), with each asked to describe 5 of the 25 stimuli. In total, 100 conversations were recorded: 44 intersections, 19 bypassing, and 37 pedestrian scenarios.

\begin{table}[h!]
\centering
\caption{Example conversation from human trials (\S\ref{sec:human_evaluation}). Description in red signified that it was generated using previous description and the user's feedback using \(\mathtt{SUMM}\) prompt (\S\ref{sec:conversation})} 
\label{tab:conversation_examples}
\resizebox{\textwidth}{!}{%
\begin{tabular}{@{}clll@{}}
\toprule
\multicolumn{4}{c}{\includegraphics[width=0.3\columnwidth]{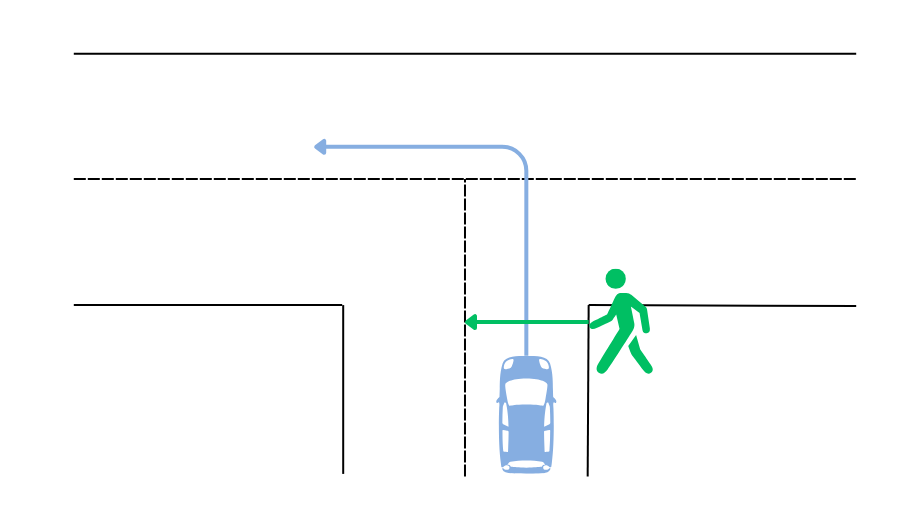}}                                                                                                                                                                                                                                                                                 \\ \midrule
\multicolumn{1}{c}{Turn} & \multicolumn{1}{c}{Feedback \(f\)}                                                                                   & \multicolumn{1}{c}{Description \(d\)}                                                                                                                                     & \multicolumn{1}{c}{Simulation} \\ \midrule
\textbf{1}               & \multicolumn{1}{c}{-}                                                                                              & \textit{\begin{tabular}[c]{@{}l@{}}at a 3-way intersection, a pedestrian crosses \\ the road. An ego vehicle waits to make a left turn.\end{tabular}}               & \begin{minipage}{.15\textwidth} \includegraphics[width=50pt]{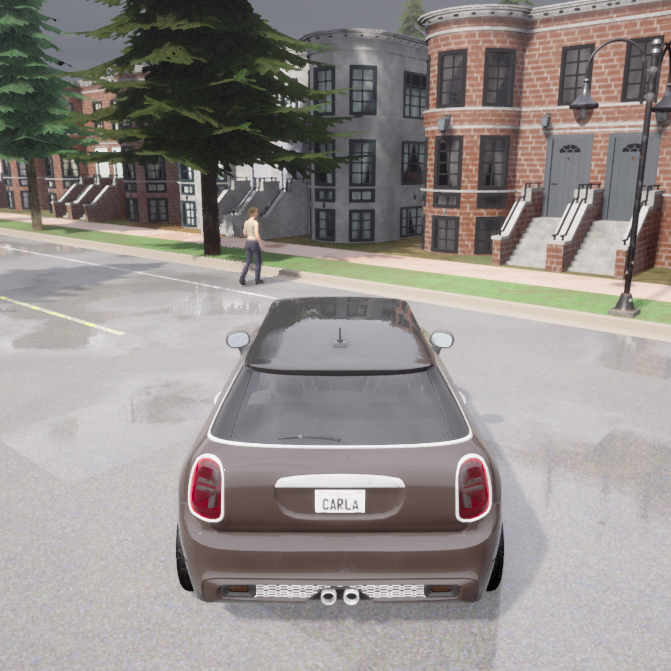} \end{minipage} \vspace{3pt}\\
\textbf{2}               & \textit{\begin{tabular}[c]{@{}l@{}}first the pedestrian crosses, \\ then ego vehicle moves.\end{tabular}}          & \textit{\color{red}\begin{tabular}[c]{@{}l@{}}At a 3-way intersection, a pedestrian crosses \\ the road and an ego vehicle waits to make \\ a left turn.\end{tabular}}            & \begin{minipage}{.15\textwidth} \includegraphics[width=50pt]{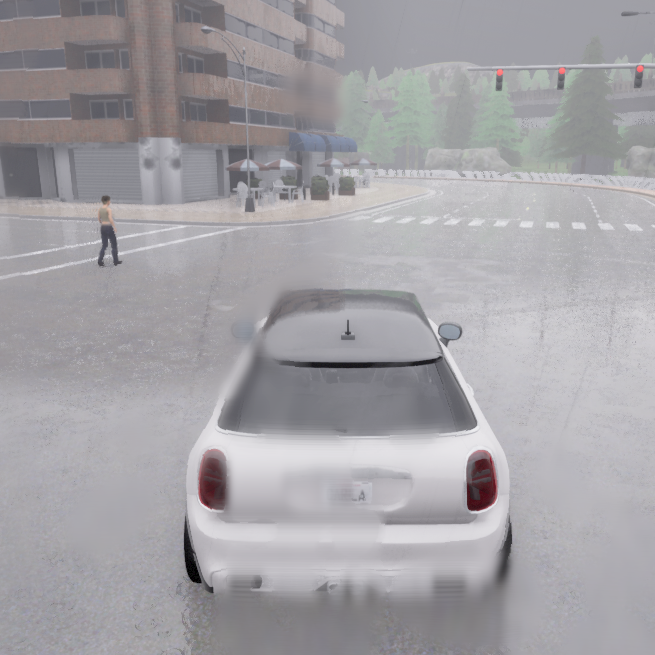}\end{minipage} \vspace{3pt}\\
\textbf{3}               & \textit{\begin{tabular}[c]{@{}l@{}}the pedestrian crosses the \\ road on which the ego \\ vehicle enters\end{tabular}} & \textit{\color{red}\begin{tabular}[c]{@{}l@{}}At a 3-way intersection, an ego vehicle waits \\ to make a left turn, and a pedestrian \\ crosses the road.\end{tabular}}             & \begin{minipage}{.15\textwidth}\includegraphics[width=50pt]{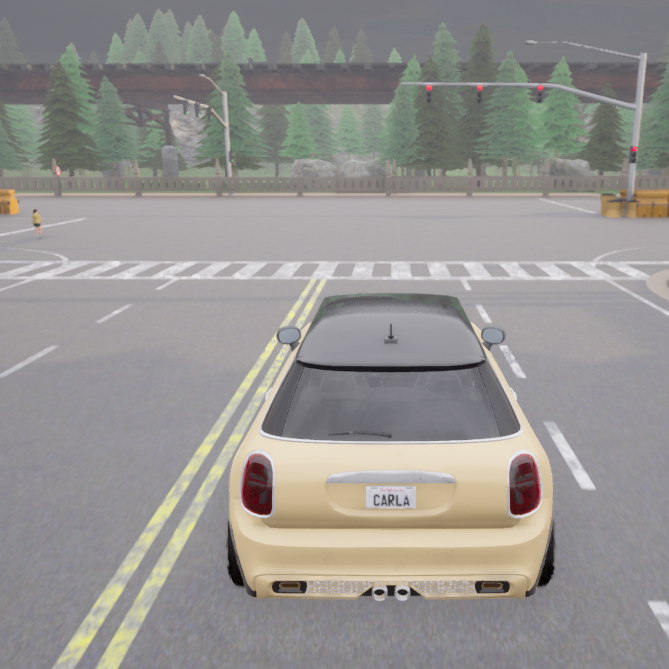} \end{minipage}  \vspace{3pt}\\
\textbf{4}               & \textit{\begin{tabular}[c]{@{}l@{}}the pedestrian is to the \\ right of the vehicle\end{tabular}}                  & \textit{\color{red}\begin{tabular}[c]{@{}l@{}}At a 3-way intersection, an ego vehicle waits to\\  make a left turn, and a pedestrian crosses \\ the road to the right.\end{tabular}} & \begin{minipage}{.15\textwidth}\includegraphics[width=50pt]{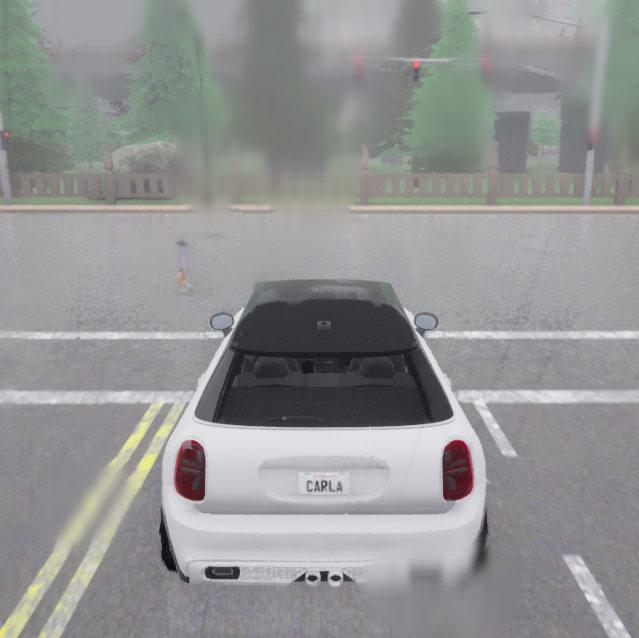}\end{minipage}  \\
\multicolumn{4}{l}{\textbf{Outcome}: simulation generated successfully}  \\ \midrule
\multicolumn{4}{c}{\includegraphics[width=0.3\columnwidth]{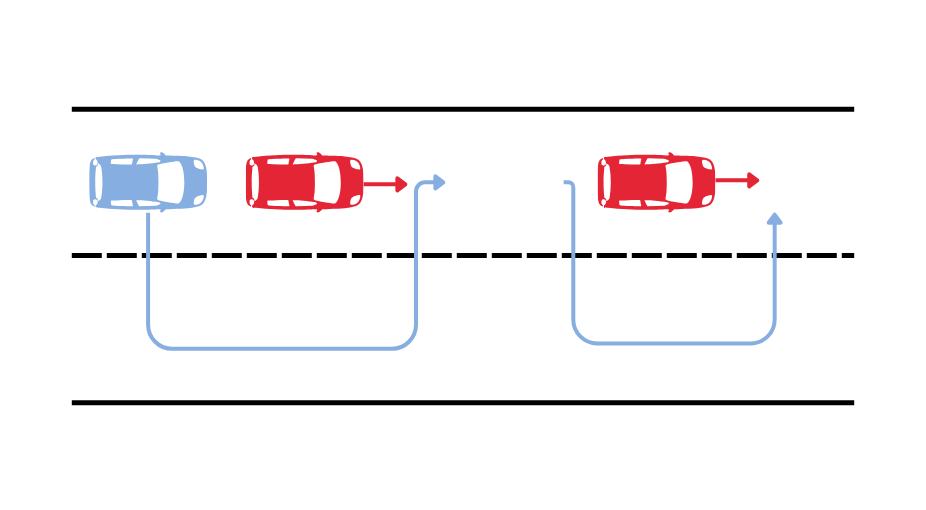}}                                                                                                                                                                                                                                                                                 \\ \midrule
\multicolumn{1}{c}{Turn} & \multicolumn{1}{c}{Feedback \(f\)}                                                                                   & \multicolumn{1}{c}{Description \(d\)}                                                                                                                                     & \multicolumn{1}{c}{Simulation} \\ \midrule
\textbf{1}               & \multicolumn{1}{c}{-}                                                                                              & \textit{\begin{tabular}[c]{@{}l@{}} An ego vehicle overtakes an adversary vehicle, \\ then overtakes another adversary vehicle\end{tabular}}               & \begin{minipage}{.15\textwidth} \includegraphics[width=50pt]{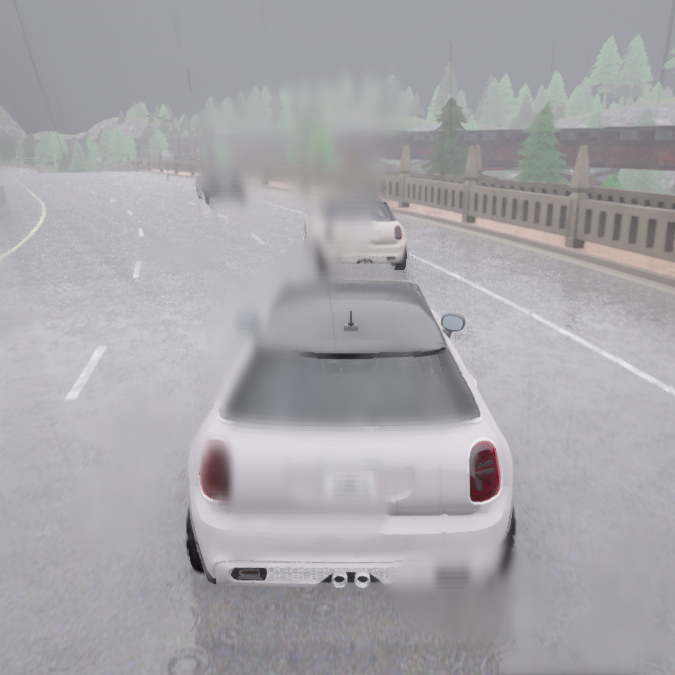} \end{minipage} \vspace{3pt}\\
\textbf{2}               & \textit{\begin{tabular}[c]{@{}l@{}}The ego vehicle is on a dual \\ carriageway behind two \\ adversary vehicles in the left lane\end{tabular}}          & \textit{\color{red}\begin{tabular}[c]{@{}l@{}}An ego vehicle overtakes an  adversary vehicle, \\ then overtakes another adversary vehicle on a \\ dual carriageway behind two adversary vehicles \\ in the left lane.\end{tabular}}            & \begin{minipage}{.15\textwidth} \includegraphics[width=50pt]{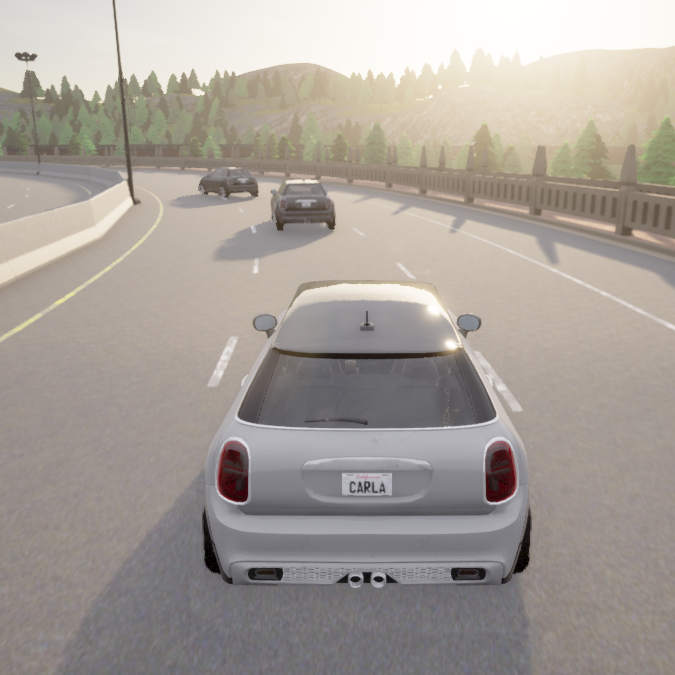}\end{minipage} \vspace{3pt}\\
\textbf{3}               & \textit{\begin{tabular}[c]{@{}l@{}}The road has two lanes and all \\ vehicles begin in the left lane\end{tabular}} & \textit{\color{red}\begin{tabular}[c]{@{}l@{}}An ego vehicle overtakes an \\ adversary vehicle, then overtakes another adversary \\ vehicle on a dual carriageway behind \\ two adversary vehicles in the left lane.\end{tabular}}             & \begin{minipage}{.15\textwidth}\includegraphics[width=50pt]{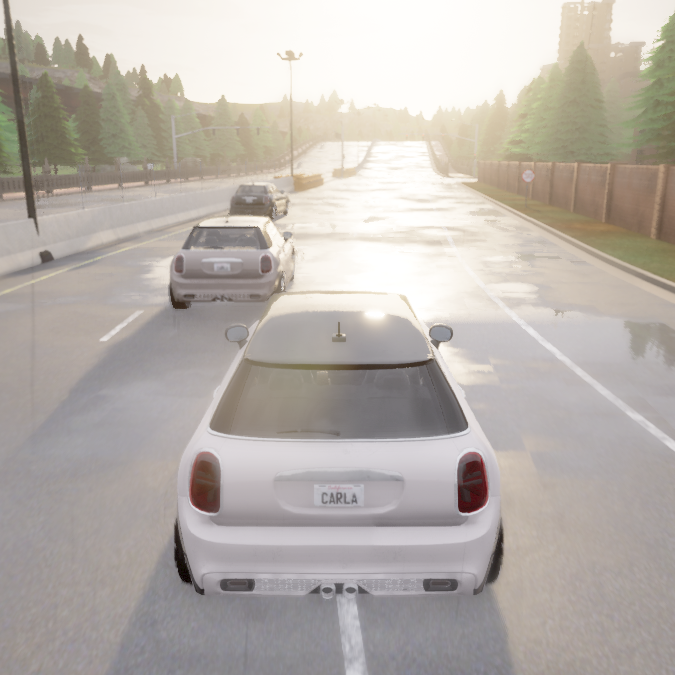} \end{minipage}  \vspace{3pt}\\
\textbf{4}               & \textit{\begin{tabular}[c]{@{}l@{}}The road only has two lanes\end{tabular}}                  & \textit{\color{red}\begin{tabular}[c]{@{}l@{}}An ego vehicle overtakes an adversary vehicle, \\ then overtakes another adversary vehicle on a \\ dual carriageway behind two adversary vehicles \\ in the left lane.\end{tabular}} & \begin{minipage}{.15\textwidth}\includegraphics[width=50pt]{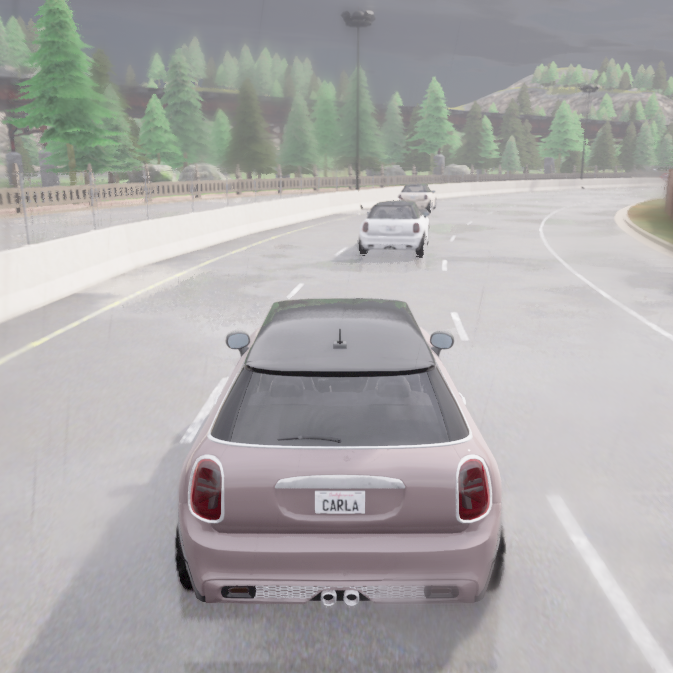}\end{minipage} \\
\multicolumn{4}{l}{\textbf{Outcome}: failed to generate the simulation}   \\ \bottomrule
\end{tabular}
}
\end{table}

\subsubsection{Results and Discussion}

Figure~\ref{fig:results} outlines the results of human evaluation. As the number of dialogue turns increases, so does the percentage of successfully generated scenarios: without dialogue, only 6\% of the code generations were successful, while conversations of up to four turns raise this to 28\%. This holds per scenario type, with or without execution errors. This suggests that the \textit{extended conversation enables successful scenario generation}. Table~\ref{tab:conversation_examples} gives examples of a successful and failed scenario generation. Error analysis shows that failed scenario generation conversations had one or more of these errors:
\begin{itemize}
    \item Execution error (62.5\%): the code produced does not execute, and the user is asked to paraphrase one or more times. This happens due to a missing correspondence between Scenic API and the description. E.g. the description features ``T-junction'', while the Scenic API would expect 3-way intersection, and the code features \texttt{junction} as a primitive rather than \texttt{intersection}.
    \item Pass error (82\%): the code executes but does not pass as a correct simulation. It is because the user's description diverges from descriptions in the dataset: in length (training data descriptions were 30-50 tokens while user descriptions were up to 200 tokens); in syntax (training data descriptions were single sentences while users expressed multiple sentences and fragments); and in narrative style (training data descriptions were ego-centric while users expressed other perspectives).
    \item Summarization error (27.8\%): errors arising due to summarization (\S\ref{sec:conversation}) like failure to include the feedback in the description or the updated description has extra information.
\end{itemize}

\section{Conclusions}
\label{sec:conclusions}

We have presented a dialogue system for generating conversational driving scenarios. Our results show that users can, in principle, generate desired driving scenarios with conversation being a crucial element in reducing scenario generation errors. 

For future work, we envision several developments. Firstly, there is a gap in generating scenarios, despite the extended conversation. To address this, further techniques could be considered, including data augmentation \cite{chen-lampouras-2023-exploring}, constrained decoding \cite{geng-etal-2023-grammar}, prompt engineering \cite{DBLP:journals/corr/abs-2307-08220}, and memory-based techniques for explicit state tracking \cite{jain-lapata-2021-memory}. Second, the current setup does not dynamically change the controller and use the interface to test it. This should be evaluated and considered testing of such new controllers in tandem with formal verification methods \cite{DBLP:conf/seke/Wang0SO23}.

\begin{acknowledgments}
  This work was supported by UKRI Strategic Priorities Fund to the UKRI Research Node on Trustworthy Autonomous Systems Governance and Regulation (grant EP/V026607/1) and  UKRI Turing AI World Leading Researcher Fellowship on AI for Person-Centred and Teachable Autonomy (grant EP/Z534833/1)
\end{acknowledgments}

\section*{Declaration on Generative AI}

During the preparation of this work, the author(s) used Grammarly in order to do the following: Grammar and spell checking; generating a paraphrase or alterantive word. After using this tool, the author(s) reviewed and edited the content as needed, and we take full responsibility for the publication’s content.


\end{document}